%% file: ms.tex
\documentclass[10pt,twocolumn,letterpaper]{article}

\usepackage{wacv}              %

\usepackage{graphicx}
\usepackage{amsmath}
\usepackage{amssymb}
\usepackage{booktabs}
\usepackage{enumitem}%
\usepackage[hang,flushmargin]{footmisc}
\usepackage{placeins}
\usepackage[accsupp]{axessibility}

\usepackage[pagebackref,breaklinks,colorlinks]{hyperref}

\usepackage[capitalize]{cleveref}
\crefname{section}{Sec.}{Secs.}
\Crefname{section}{Section}{Sections}
\Crefname{table}{Table}{Tables}
\crefname{table}{Tab.}{Tabs.}

\newif\ifarxiv
\arxivtrue

\def\wacvPaperID{1796} %
\def\confName{WACV}
\def\confYear{2025}

\input{preamble}

\begin{document}
\twocolumn

\title{Swap Path Network for Robust Person Search Pre-training}

\ifarxiv
    \author{Lucas Jaffe\textsuperscript{1,2}\thanks{Correspondence to: Lucas Jaffe $<$jaffe5@llnl.gov$>$} \\
        \textsuperscript{1}Lawrence Livermore National Laboratory\\
        \and
        Avideh Zakhor\textsuperscript{2}\\
        \textsuperscript{2}University of California, Berkeley\\
    }
\else
    \author{Lucas Jaffe\textsuperscript{1,2}\\
        \textsuperscript{1}Lawrence Livermore National Laboratory\\
        {\tt\small jaffe5@llnl.gov}
        \and
        Avideh Zakhor\textsuperscript{2}\\
        \textsuperscript{2}University of California, Berkeley\\
        {\tt\small avz@berkeley.edu}
    }
\fi

\maketitle

\input{sec/0_abstract}
\input{sec/1_intro}
\input{sec/2_relatedwork}

\input{sec/3_methods}
\input{sec/4_results}

\input{sec/5_conclusion}

\input{sec/6_acknowledgements}

\ifarxiv
    {\small
    \bibliographystyle{ieee_fullname}
    \bibliography{ms}
    }
\else
    {\small
    \bibliographystyle{ieee_fullname}
    \bibliography{main}
    }
\fi

\onecolumn
\input{sec/X_suppl}
\ifarxiv
\else
\input{sec/6_acknowledgements}
\fi

\end{document}

%% file: preamble.tex
\usepackage[dvipsnames]{xcolor}

\DeclareMathOperator{\csim}{sim}
\DeclareRobustCommand{\rchi}{{\mathpalette\irchi\relax}}
\newcommand{\irchi}[2]{\raisebox{\depth}{$#1\chi$}} %
\usepackage{multirow}
\newcommand{\num}[2]{${#1}\mathrm{e}{#2}$}
\usepackage{caption} 
\usepackage{float}
\usepackage{afterpage}
\usepackage{wrapfig}

\let\titleold\title
\renewcommand{\title}[1]{\titleold{#1}\newcommand{\thetitle}{#1}}
\def\maketitlesupplementary
   {
   \newpage
        {\centering
        \Large
        \textbf{\thetitle}\\
        \vspace{0.5em}Supplementary Material \\
        \vspace{1.0em}}
   }

%% file: sec/0_abstract.tex
\begin{abstract}
In person search, we detect and rank matches to a query person image within a set of gallery scenes. Most person search models make use of a feature extraction backbone, followed by separate heads for detection and re-identification. While pre-training methods for vision backbones are well-established, pre-training additional modules for the person search task has not been previously examined. In this work, we present the first framework for end-to-end person search pre-training. Our framework splits person search into object-centric and query-centric methodologies, and we show that the query-centric framing is robust to label noise, and trainable using only weakly-labeled person bounding boxes. Further, we provide a novel model dubbed Swap Path Net (SPNet) which implements both query-centric and object-centric training objectives, and can swap between the two while using the same weights. Using SPNet, we show that query-centric pre-training, followed by object-centric fine-tuning, achieves state-of-the-art results on the standard PRW and CUHK-SYSU person search benchmarks, with 96.4\% mAP on CUHK-SYSU and 61.2\% mAP on PRW. In addition, we show that our method is more effective, efficient, and robust for person search pre-training than recent backbone-only pre-training alternatives.
\end{abstract}

%% file: sec/1_intro.tex
\section{Introduction}
\label{sec:intro}

\textit{Person search} is the combined formulation of two subproblems:
\textit{detection} of all person bounding boxes in a set of gallery scenes, and \textit{re-identification} (\textit{re-id}) of all detected boxes with respect to a query person box.
Recent person search models are mainly \textit{end-to-end}, and have a feature extraction backbone, followed by separate heads for detection and re-identification.
It is typical to initialize this feature backbone using weights pre-trained for classification on the \textit{ImageNet-1k} dataset \cite{russakovsky_imagenet_2015}.
Only recently have approaches emerged considering pre-training backbones for person search on annotated person data \cite{chen_beyond_2023, jia_collaborative_2023}, even then only considering cropped bounding boxes from the \textit{LUPerson} \cite{fu_unsupervised_2021} or \textit{Market} \cite{zheng_scalable_2015} datasets.

\begin{figure}
  \centering
  \includegraphics[width=\linewidth]{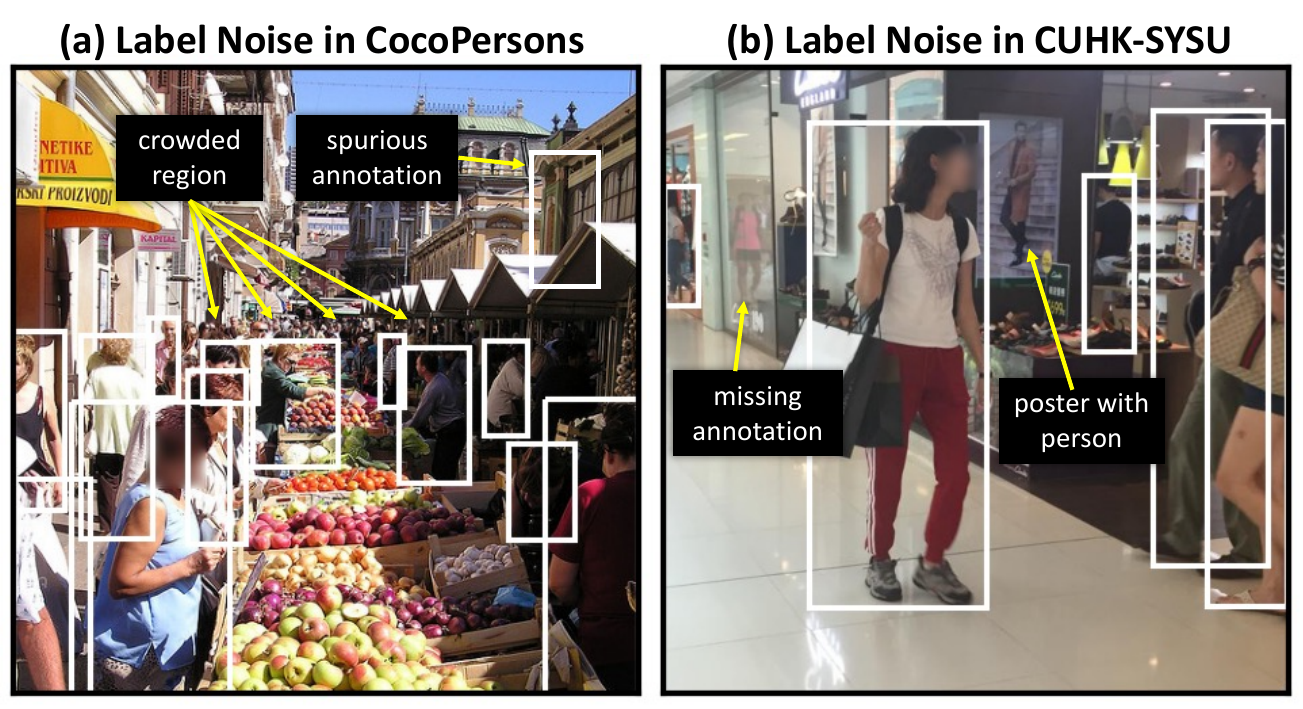}
  \vspace{-0.7cm}
  \caption{Examples of label noise and annotation challenges in real annotated scenes, with white boxes used for annotations.
    }
  \vspace{-0.5cm}
  \label{fig:label_noise}
  \centering
\end{figure}

For the detection portion of person search, annotations and predictions typically take the form of rectangular bounding boxes.
In datasets annotated for person detection, label noise is unavoidable, shown in \cref{fig:label_noise}.
For example, how small does a person have to be before they become part of a crowd, shown in \cref{fig:label_noise}a?
How does a model handle the presence of missing annotations or images of people in the scene, shown in \cref{fig:label_noise}b?
What if a person is behind a window in a building, visible in a monitor screen, or reflected in a mirror?
If a detector is used to automatically label people in a scene, how does a model handle spurious extra annotations?
These problems are compounded by pre-training, because annotation biases in the pre-training dataset may not reflect those in the target fine-tuning dataset.

This leaves three key issues unresolved in the pursuit of effective pre-training for person search:
1) current backbone pre-training is done using a pretext task unrelated to the person search task,
2) all parameters beyond the backbone are randomly initialized, and
3) current pre-training approaches do not consider robustness to label noise and annotation biases present in person data.
Therefore, a model is needed which can simultaneously pre-train for detection and re-id, while achieving domain transfer from a pre-training dataset of heterogeneous imagery that contains label noise and bias.

\begin{figure}
  \centering
  \includegraphics[width=\linewidth]{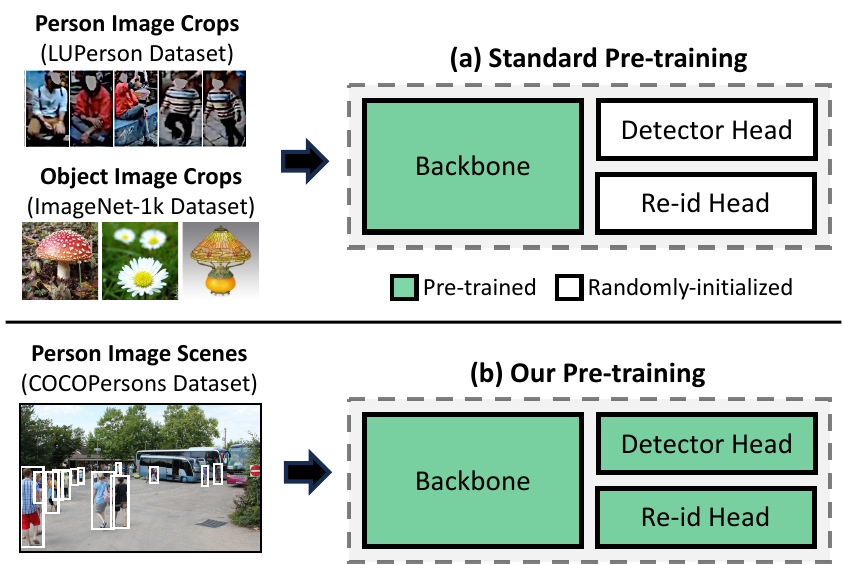}
  \caption{The standard person search model pre-training approach (shown top) pre-trains only backbone weights using either ImageNet-1k classification or person image crops from \eg, LUPerson. Our approach (shown bottom) initializes all model weights using full person scenes with multiple annotated persons. (LUPerson images from paper \cite{fu_unsupervised_2021})}
  \vspace{-0.3cm}
  \label{fig:pretraining}
  \centering
\end{figure}

In this paper, we develop a novel model which addresses these challenges by pre-training using the \textit{query-centric detection} pretext task.
In query-centric detection, we detect matches to a query object in a gallery scene, vs. standard \textit{object-centric detection}, where we detect a pre-determined set of objects in a scene.
The model is named \textbf{S}wap \textbf{P}ath \textbf{Net} (\textbf{SPNet}) because it can swap between query-centric and object-centric pathways, while using the same weights.
The two pathways are needed because the query-centric mode is robust to label noise and learns more generalizable features, making it ideal for pre-training, while the object-centric mode performs better at person search and is much more efficient during inference, making it preferable for fine-tuning on person search.
In addition, SPNet is capable of pre-training using \textit{weakly-labeled} person bounding boxes, \ie identity correspondence between scenes is not known.
We visualize our pre-training approach vs. typical backbone-only pre-training in \cref{fig:pretraining}.

We show that when SPNet is pre-trained in the query-centric mode, and fine-tuned in the object-centric mode, it achieves state-of-the-art performance on the benchmark \textit{CUHK-SYSU} \cite{xiao_joint_2017} and \textit{Person Re-identification in the Wild} (\textit{PRW}) \cite{zheng_person_2017} person search datasets.
We demonstrate that weakly-supervised pre-training on the \textit{COCOPersons} dataset \cite{lin_microsoft_2014, shao_crowdhuman_2018} using our method is more effective, efficient, and robust than recent backbone-only pre-training alternatives for unsupervised person re-id \cite{chen_beyond_2023, fu_large-scale_2022, zhu_pass_2022}.

Our contributions are as follows:
\begin{itemize}[noitemsep, topsep=0pt]
\item The \textit{Swap Path Network}: an efficient end-to-end model of person search which can operate in query-centric or object-centric modes.
\item A query-centric pre-training algorithm unique to the Swap Path Network that results in SOTA performance and is robust to label noise.
\end{itemize}

We support these claims with extensive experiments demonstrating the efficiency of the SPNet model and the efficacy of the pre-training approach.
Further, we ensure reproducibility by providing the code and installation instructions required to repeat all experiments, which are included in the corresponding GitHub repository\footnote{Project repository: \url{https://github.com/LLNL/spnet}}.

%% file: sec/2_relatedwork.tex
\section{Related Work}
\label{sec:related_work}

\begin{figure*}
  \centering
  \includegraphics[width=\linewidth]{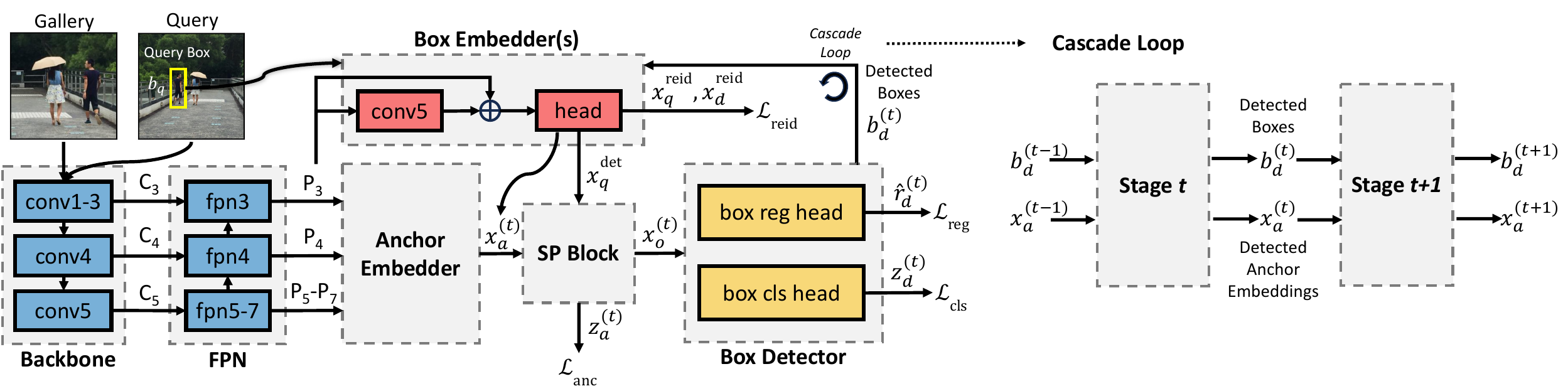}
  \caption{
Full SPNet architecture is shown on the left, with details about the Cascade Loop shown on the right.
The subscripts $q, d, a$ stand for ``query'', ``detection'', and ``anchor'' respectively.
The superscript ``reid'' means the embedding or loss is used for re-id, and the superscript ``det'' means the embedding is used for detection.
}
  \vspace{-0.3cm}
  \label{fig:spnet}
  \centering
\end{figure*}

\subsection{Person Search}

{\noindent {\bf Weakly-Supervised.}} In the context of person search, weak supervision (WS) refers to training on person bounding boxes without identity-level labels.
Several methods \cite{jia_collaborative_2023,han_context-aware_2021,wang_deep_2023,yan_exploring_2022,wang_self-similarity_2023,han_weakly_2021} have emerged in recent years to tackle this problem by using weakly-supervised objectives directly on the target dataset.
These methods focus on clustering visual features and applying contextual information to determine pseudo-labels used for a contrastive re-id loss.
In contrast, our method is orthogonal to the problem of determining pseudo-labels, focusing on the query-centric vs. object-centric training distinction.
In addition, we apply weak supervision only during pre-training on a source dataset, then perform fully-supervised fine-tuning on the target dataset.
We note that other methods assume the underlying data has multiple images per identity, and exploit this to form pseudo-labels based on common features, while we do not.
Therefore, other methods may perform better for the WS scenario on PRW, CUHK-SYSU, but are not suitable for pre-training on large unlabeled person datasets where there are few or only one image per person like COCOPersons.

{\noindent {\bf Query-Centric vs. Object-Centric.}}
Most person search models implement an \textit{object-centric} (OC) approach \cite{dong_leaps_2023,cao_pstr_2022,yu_cascade_2022,li_sequential_2021,li_cross-scale_2021,yan_anchor-free_2021}, in which we detect persons independent of any query, then afterwards compare queries to detected persons. By contrast, \textit{query-centric} (QC) person search models \cite{munjal_query-guided_2019,munjal_query-guided_2023,dong_instance_2020,wang_tcts_2020} use query person images during the detection process, producing proposals tailored to each query, but significantly increasing time complexity of inference computation.

Despite the greater time complexity of query-centric person search, multiple successful approaches have emerged.
QEEPS \cite{munjal_query-guided_2019} and QGN \cite{munjal_query-guided_2023} use query information both during detection and computation of re-id embeddings by combining query features with backbone features using squeeze-and-excitation layers \cite{hu_squeeze-and-excitation_2018}. IGPN \cite{dong_instance_2020} and TCTS \cite{wang_tcts_2020} are two-step models which extract person embeddings from a separate network, and combine these embeddings with features from the detector network. In our approach, we consider query-centric (QC) learning as a pre-training objective, while fine-tuning in the object-centric (OC) mode, though our framework supports query-centric evaluation as well. This allows us to learn more robust features in the QC mode, while retaining efficient evaluation in the OC mode.

{\noindent {\bf Pre-training.}} Person search models typically use backbone weights initialized from standard ImageNet-1k classifier training, while other model weights are initialized randomly.
To our knowledge, the only other work to date to explore another backbone initialization strategy is SOLIDER \cite{chen_beyond_2023}.
The SOLIDER framework trains on unlabeled person crops from the LUPerson dataset \cite{fu_unsupervised_2021} by optimizing backbone features using cluster pseudo-labels based on tokens learned with DINO \cite{caron_emerging_2021}.
Unlike our proposed method, SOLIDER initializes only the backbone, and not the rest of the model. In addition, SOLIDER trains only on person crops, while we train on full person scenes, containing potentially many persons.
While SOLIDER is the only prior pre-training method which measures fine-tuning for the person search task, there is a rich body of work on similar unsupervised person re-id methods \cite{fu_unsupervised_2021-1, fu_large-scale_2022, luo_self-supervised_2021, cho_part-based_2022, yang_unleashing_2022, zhu_pass_2022}, which also train only on cropped person images.

\subsection{Unsupervised Detector Pre-training}

In UP-DETR \cite{dai_up-detr_2021}, a DETR detector model \cite{carion_end--end_2020} is pre-trained with the \textit{random query patch detection} pretext task, and fine-tuned for object detection. This model is the closest current model to ours in function, as it supports query-centric detection pre-training with random patches, and object-centric fine-tuning without layer modifications. By comparison, our model implements search instead of only detection, and optimizes much faster due to the explicit spatial inductive bias of anchor-based models vs. DETR. In DETReg \cite{bar_detreg_2022}, a DETR model is trained with pseudo-label boxes generated by selective search \cite{uijlings_selective_2013}. While this approach has the potential to transfer well if the selective search object distribution resembles that of the downstream task, this is unlikely in practice, and the model will have to ``unlearn'' bad pseudo-boxes during fine-tuning, an issue which is avoided with our query-centric approach.

%% file: sec/3_methods.tex
\section{Methods}

\subsection{SPNet Description}
\label{sec:model}

A diagram detailing components of SPNet and the Cascade Loop subcomponent is shown in \cref{fig:spnet}.
SPNet takes Gallery images as input, and outputs detected boxes $b_d$ with corresponding class logits $z_d$ and re-id embeddings $x_d^\text{reid}$.
SPNet also takes a Query scene as input with a corresponding ground truth person box $b_q$, and outputs the re-id embedding for that box $x_q^\text{reid}$.
The query re-id embedding $x_q^\text{reid}$ is compared to gallery re-id embeddings $x_d^\text{reid}$ via cosine similarity to produce re-id scores, which are used to rank predictions for retrieval.
Image features (C$_3$-C$_5$) are learned by the backbone and refined by the FPN (P$_3$-C$_7$), and then fed to the Box Embedder, with either query boxes $b_q$ or detected boxes $b_d$, to learn re-id embeddings $x^\text{reid}, x^\text{det}$.
FPN features are also fed to the Anchor Embedder to learn initial anchor embeddings $x_a^0$.

\begin{figure*}
  \centering
  \includegraphics[width=\linewidth]{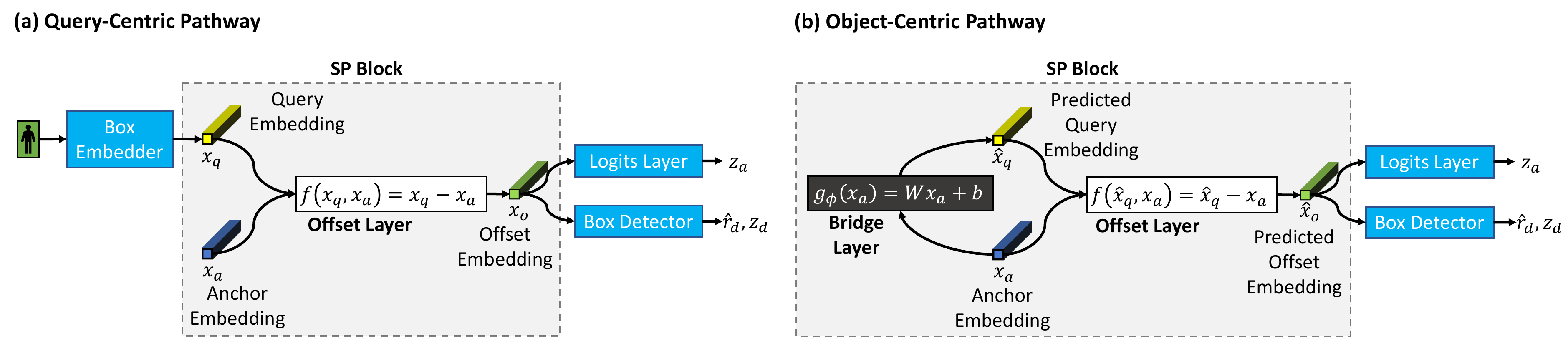}
  \caption{Query-centric (a) and object-centric (b) pathways of the SP Block.
    Note that the query-centric pathway takes as input a query embedding $x_q$ extracted from a person image, while the object-centric pathway predicts the query embedding $\hat{x}_q$ directly from a matching anchor embedding $x_a$ using the Bridge Layer $g_\phi$.
}
  \vspace{-0.3cm}
  \label{fig:qc_oc}
  \centering
\end{figure*}

The Box Embedder can be optionally duplicated to produce separate query embeddings for re-id $x_q^\text{reid}$ and detection $x_q^\text{det}$, or shared \ie $x_q^\text{reid} = x_q^\text{det}$, shorthanded as $x_q$.
Using a shared Box Embedder has a regularizing effect by pushing $x_q$ to comply with both re-id and detection losses, while duplicating the Box Embedder gives more capacity for each task, which can lead to overfitting.
Embeddings $x_a$, and $x_q$ for QC, then enter the SP Block, where they may pass through either the QC or OC pathway, depicted in \cref{fig:qc_oc}.

{\noindent {\textbf{QC Pathway.}}}
In the QC pathway, shown in \cref{fig:qc_oc}a, we compute offset embeddings $x_o$ from the \textit{Offset Layer}, by simply subtracting each anchor embedding from the query embedding: $x_o = x_q - x_a$.
Offset embeddings $x_o$ are passed to a \textit{Logits Layer}, explained in \cref{sec:losses}, to produce classifier predictions $z_a$ for each anchor box, which predict whether a given anchor matches the query $q$.
These $z_a$ are used similarly to classifier logits in a standard Faster R-CNN Regional Proposal Network (RPN) \cite{ren_faster_2015}: they filter a large number of anchors, typically more than $100,000$ down to less than $1,000$, which are then refined to produce more accurate boxes.

{\noindent {\textbf{OC Pathway.}}}
In the OC pathway, shown in \cref{fig:qc_oc}b, we do not have knowledge of any query embedding $x_q$, so we instead compute a pseudo-query embedding $\hat{x}_q$ using the \textit{Bridge Layer}, by default a single affine layer $g_\phi(x_a)=Wx_a + b$ with learnable parameters $\phi=\{W, b\}$.
By mimicking the QC pathway instead of predicting localization offsets and logits directly from $x_a$, we improve transfer performance between the two pathways.
Crucially, the layer $g_\phi$ is the only difference between the QC and OC models, meaning that all SPNet weights, aside from $\phi$, can be trained in one mode, and ported to the other, or vice versa.

{\noindent {\bf Box Prediction.}}
Both pathways output offset embeddings $x_o$ that are passed to a separate Box Detector module, shown in \cref{fig:spnet}, which performs the box refinement. The \texttt{box reg head} is a 4-layer MLP with input $x_o$ which predicts offsets $\hat{r}_d$ between anchor boxes and matching ground truth boxes. The \texttt{box cls head} uses the same Logits Layer as the SP Block to produce class logits $z_d$.
Predicted boxes $b_d$ are fed back to the Box Embedder, the same as that used to produce $x_q^\text{reid}$, to produce predicted re-id embeddings $x_d^\text{reid}$.
The $x_q^\text{reid}$ from queries are compared to $x_d^\text{reid}$ from gallery images for ranking.

{\noindent {\bf Cascade Loop.}}
We can iteratively refine box accuracy and scores by repeating the same basic prediction structure in a loop, referred to as the \textit{Cascade Loop} in \cref{fig:spnet}, originating from the Cascade R-CNN detector \cite{cai_cascade_2018}, and also done in the SeqNet \cite{li_sequential_2021} and SeqNeXt \cite{jaffe_gallery_2023} person search models.
In the Cascade Loop, predicted boxes $b_d^{(t)}$ from stage $t$ are fed back into the Box Embedder to produce anchor embeddings $x_a^{(t)}$ for the next step $t+1$.
The outputs $z_a^{(t)}$, $x_o^{(t)}$, $z_d^{(t)}$, and $\hat{r}_d^{(t)}$ are also updated during each Cascade Loop step.
The detector re-id embeddings $x_d^\text{reid}$ are computed from the detected boxes coming out of the final Cascade Loop stage.
The SP Block, Box Detector, and Box Embedder are all duplicated for each round, meaning that we use modules with the same architecture, but do not share weights of the same module between rounds.

\subsection{Pretext Task}
\label{sec:pretext_task}

\begin{figure*}
  \centering
  \includegraphics[width=\linewidth]{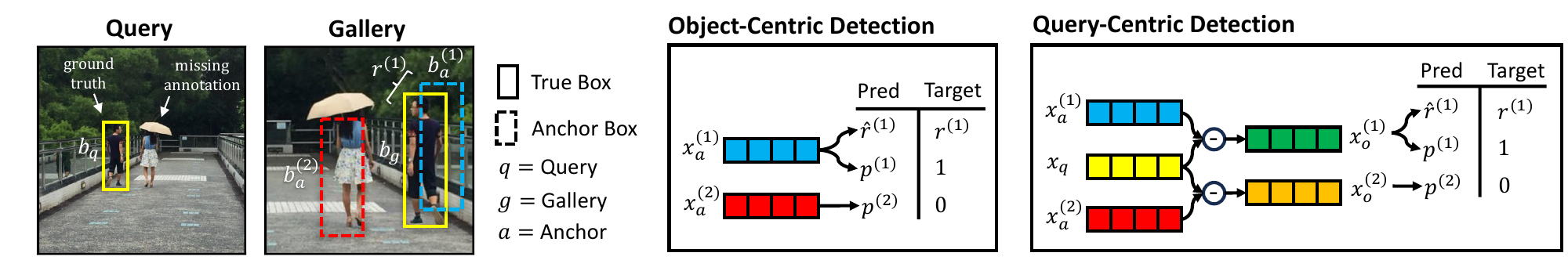}
  \caption{Visual comparison of object-centric (OC) and query-centric (QC) detection tasks between two augmentations of the same base image (Query, Gallery). One person box is annotated (ground truth $b_q$, $b_g$), while the other is not (missing annotation). Note that anchor box$_1$ $b_a^{(1)}$ overlaps ground truth box $b_g$ while anchor box$_2$ $b_a^{(2)}$ does not. Anchor embeddings $x_a$ are used to compute box offsets $\hat{r}$ and anchor probabilities $p$ either directly (OC) or relatively using $x_o = x_q - x_a$ (QC). We do not compute box offsets $\hat{r}$ for $b_a^{(2)}$ because it does not match any ground truth.}
  \vspace{-0.3cm}
  \label{fig:pretext_task}
\end{figure*}

The goal of our pre-training pretext task is to initialize nearly all model weights to optimize for transfer to the downstream person search task. In \cref{fig:pretext_task}, we show the QC detection pretext task compared to standard OC detection.
In OC detection, the goal is to regress each anchor class probability $p$ to 1 if the anchor matches a ground truth box $b_g$ \eg, $b_a^{(1)}$, or 0 if it does not \eg, $b_a^{(2)}$.
The overlap threshold for matching boxes is $0.5$ IoU.
For matching anchors, we also regress box offsets $\hat{r}^{(1)}$ to targets $r^{(1)}$.
In QC detection, the regression targets are the same, but predictions for each anchor are computed only \textit{relative} to a query, explained by the QC vs. OC pathways in the previous section.
QC detection results in more robust optimization, because it can better handle label noise like the missing annotation in \cref{fig:pretext_task} or examples in \cref{fig:label_noise}.
Intuitively, QC detection allows us to learn salient features for transfer to person search without rigidly defining positive and negative detections.

Our pre-training experiments utilize weakly-labeled person boxes.
Since identities are not tracked between images for weakly-labeled data, we generate correspondence between images using data augmentation: we know that a box in an image corresponds to the same box in an augmented version of that image, with potentially different coordinates, visualized in \cref{fig:pretext_task}.
Augmentations consist of scaling, cropping, and horizontal flipping.

For QC detection on weakly-labeled data, each image in a batch is augmented $k$ times, where $k=2$ for all experiments, with each augmented image taking a turn as both query and gallery. Annotated boxes in the query image are compared against the corresponding box in the gallery image, and assigned to anchors with overlap $\geq 0.5$ IoU. Self comparisons are allowed as well, in which the query and gallery are the same image.

For all cases, OC optimization treats annotated boxes as true objects with the same anchor matching criterion as QC, and pushes anchor scores to 0 or 1 as shown in \cref{fig:pretext_task}. We also augment batch images $k$ times for the OC case to fairly compare with QC, though it is not necessary.

\subsection{Losses}
\label{sec:losses}

{\noindent {\bf Classifier Losses.}}
For the offset embeddings $x_o$ shown in \cref{fig:spnet}, it is critical to define a loss function which enforces a consistent relationship between query embeddings $x_q$ and anchor embeddings $x_a$, while performing the classification regression. The goal is to regress query-anchor matches to 1 and non-matches to 0, shown in \cref{fig:pretext_task}. For this task, we model the anchor loss after the Norm-Aware Embedding concept from \cite{chen_norm-aware_2020}, and the Focal Loss from \cite{lin_focal_2017}.

Recall the offset embedding is defined as the difference between query and anchor embeddings: $x_o = x_q - x_a$ with $x_o \in \mathbb{R}^d$. Assign $w = \| x_o \|$. In the Norm-Aware Embedding work \cite{chen_norm-aware_2020}, classification logits are computed from embedding norms using a Batch Norm layer \cite{ioffe_batch_2015}.%
To avoid use of unstable batch statistics, we instead compute a fixed rescaling by modeling $x_o \sim \mathcal{N}(0, I_d)$, which implies $w \sim \textrm{Chi}(d)$.
Then we can standardize $w$ using the mean and standard deviation of the Chi distribution, with class logits given by $z=-(w - \mu_{\text{Chi}}) / s_{\text{Chi}}$.
Finally, we compute class probabilities with the logistic sigmoid $p = \sigma(z)$.

For anchor logits $z_a$, we compute a loss averaged across all anchors $\mathcal{L}_\text{anc}$, applying the Focal Loss defined in \cite{lin_focal_2017}, with hyperparameters $\alpha_\text{\tiny{FL}} = 0.5, \gamma_\text{\tiny{FL}} = 1$. For cascade layer class logits $z_d$, we compute a loss $\mathcal{L}_\text{cls}$ averaged across predicted boxes using the unweighted Cross Entropy Loss.

Additional justification is given in Supplementary \cref{supp:sec:model_arch_ablation}, where we compare our \textit{FixedNorm} logits formulation vs. the standard BatchNorm, and describe other intuitive benefits of our formulation.

{\noindent {\bf Box Regression Loss.}}
For bounding box offset regression, we use the generalized IoU (GIoU) loss \cite{rezatofighi_generalized_2019}, shown as $\mathcal{L}_\text{reg}$. This loss has a beneficial interaction with the classifier loss: for matching query and anchor boxes (high IoU/GIoU), the target box regression offsets should be small, corresponding to $\|x_o\|$ small. This allows the model to learn that the magnitude of $\|x_o\|$ correlates directly to the size of predicted box offsets.

{\noindent {\bf Re-id Losses.}}
For our re-id losses, we use two variations of the normalized temperature-scaled cross-entropy loss \cite{sohn_improved_2016,chen_simple_2020}. We use the terms \textit{positive pair} and \textit{negative pair} to refer to two embeddings with the same label or different labels respectively.

We define the probability for positive pair $(x, x^+)$ under the contrastive objective as
\begin{equation}
p(x, x^+) = \frac{
        s_{\tau}(x, x^+)
}{
        s_{\tau}(x, x^+) + \sum_{x^- \in \rchi_x^-}s_{\tau}(x, x^-)
}
\label{eqn:contrastive_prob}
\end{equation}

With the full contrastive loss expressed as
\begin{equation}
\mathcal{L}_\text{reid} = -\log \sum\nolimits_{x \in \rchi} \sum\nolimits_{x^+ \in \rchi_x^+} p(x, x^+)
\label{eqn:contrastive_loss}
\end{equation}

using $s_{\tau}(u, v) \triangleq \exp{(\csim(u, v)/\tau)}$ and $\csim(u, v) \triangleq u \cdot v / \|u\|\|v\| , u,v \in \mathbb{R}^d$. $\rchi$ denotes the set of all $x$, $\rchi_x^+$ the set of all positive samples for $x$, and $\rchi_x^-$ the set of all negative samples for $x$. To produce variations of this loss, we vary compositions of the sets $\rchi_x^+$ and $\rchi_x^-$.

{\noindent {\bf Fine-Tuning Re-id Loss.}} During fine-tuning, we use the standard online instance matching (OIM) loss from \cite{xiao_joint_2017}.

 {\noindent {\bf Pre-Training Re-id Loss.}} During pre-training, we use a variation of the momentum contrast loss (MoCo) from \cite{he_momentum_2020}. Like in MoCo, we define an encoder network which is updated via gradient descent, and a momentum network, which is updated as a moving average of the encoder. We call embeddings from the encoder network $x_e$ and embeddings from the momentum network $x_m$. Define $\bar{x}_m$ as the mean of all predicted box embeddings with corresponding box having IoU $\geq 0.7$ with a given ground truth box. We store embeddings $\bar{x}_m$ in a queue during training. Then, we form positive pairs $\rchi_{x_e}^+$ with all embeddings $(x_e, \bar{x}_m^+)$ and negative pairs $\rchi_{x_e}^-$ with all embeddings $(x_e, \bar{x}_m^-)$.

We are effectively comparing current embeddings to moving average cluster centroids \cite{jia_collaborative_2023}, in the limit where the cluster consists of only one person image. In absence of cluster pseudo-labels similar to those used in \cite{jia_collaborative_2023} and other weakly-supervised methods, it is critical to have a representative embedding to compare against aside from the image itself. As in \cite{dong_leaps_2023}, we found it useful to utilize proposals with IoU $\geq 0.7$, but did not find it beneficial to weight them by IoU in the re-id loss, as in that work.

{\noindent {\bf Final Loss.}} The final loss for both pre-training and fine-tuning is simply the unweighted sum of all described losses:
$\mathcal{L} = \mathcal{L}_\text{reid} + \mathcal{L}_\text{anc} + \sum_{t=1}^{T+1} \mathcal{L}_\text{reg}^{(t)} + \mathcal{L}_\text{cls}^{(t)}$, where $T$ is the number of cascade stages.

%% file: sec/4_results.tex
\section{Experiments}

\subsection{Datasets and Evaluation}

{\noindent {\bf Datasets.}}
We perform weakly-supervised pre-training on the train partition of the COCOPersons dataset \cite{shao_crowdhuman_2018}, which contains 64k scenes with 257k person box annotations. COCOPersons is the subset of MS-COCO2017 \cite{lin_microsoft_2014} images containing at least one person annotation; non-person annotations are ignored. We fine-tune models on the two standard person search benchmark datasets: CUHK-SYSU \cite{xiao_joint_2017} and PRW \cite{zheng_person_2017}. Other metadata about the datasets are presented in Supplementary \cref{supp:sec:metadata}.

{\noindent {\bf Evaluation.}}
We evaluate all models by measuring fine-tuning performance once on the standard retrieval test scenario for CUHK-SYSU or PRW, described in Supplementary \cref{supp:sec:metadata}. While we compare QC vs. OC pre-training, fine-tuning is always done in the OC mode. In Supplementary \cref{supp:sec:qc_oc_ablation}, we show that QC fine-tuning is less effective than OC fine-tuning.

To measure performance, we use standard detection performance metrics of recall@0.5 IoU and average precision@0.5 (AP@0.5) IoU, indicating predictions with an overlap of $>$ 0.5 IoU with a ground truth box are considered matches.
For person search evaluation, we use the standard metrics of mean average precision (mAP) and top-1 accuracy (top-1).

\subsection{Implementation Details}
\label{sec:imp_details}

We describe the most important implementation details, with additional information given in Supplementary \cref{supp:sec:imp_details}.

{\noindent {\bf Model Configurations.}}
We perform experiments with two variants of the SPNet model: SPNet-S(mall) and SPNet-L(arge), with differences shown in \cref{tab:config}. Unless otherwise stated, SPNet-S uses a ConvNeXt-T \cite{liu_convnet_2022} backbone, and SPNet-L uses a ConvNeXt-B backbone.

{\noindent {\bf Optimization.}}
For all experiments, we pre-train and fine-tune for 30 epochs each, using the AdamW optimizer \cite{loshchilov_decoupled_2022}, a cosine-annealed schedule, and linear warmup following \cite{loshchilov_sgdr_2016, loshchilov_decoupled_2022}.

{\noindent {\bf Pre-training Optimization.}}
For experiments pre-training SPNet with our method, the backbone is initialized using ImageNet-1k classifier weights unless otherwise stated.
For the pre-training re-id loss, we use the MoCo objective described in \cref{sec:losses} with a queue length of $65,536$, momentum of $0.9999$, and temperature $\tau = 0.1$. We use the Pre-train config from \cref{tab:config} for learning rate and weight decay settings.
We also experiment with four variations of layer freezing and learning rates for the backbone and post-backbone modules, with results shown in Supplementary \cref{supp:sec:qc_oc_ablation} \cref{plot:prw_backbone_ablation}. We note that QC outperforms OC pre-training for nearly all configurations.

{\noindent {\bf Fine-tuning Optimization.}}
For the fine-tuning re-id loss, we use the OIM objective described in \cref{sec:losses} with temperature $1/30$, momentum $0.5$, and queue length $5,000$ for CUHK and $500$ for PRW, which are the standard settings. We use the Fine-tune config from \cref{tab:config}, with image size $512 \times 512$ used for SPNet-S, and image size $1024 \times 1024$ used for SPNet-L.

{\noindent {\bf Training and Inference Speed.}}
All models were trained using a single A100 GPU with 82GB VRAM. Pre-training and fine-tuning times are shown in Supplementary \cref{supp:sec:imp_details}, with the QC pre-training taking 30 hours for SPNet-S and 46.5 hours for SPNet-L.
While the SOLIDER \cite{chen_beyond_2023} authors do not give precise training times, they train for 110 epochs on 8 V100 GPUs, and reported the training took several days.
A related approach from the LUPerson paper \cite{fu_large-scale_2022} trains for 200 epochs on 8 V100 GPUs.
Another crop-only approach callled PASS \cite{zhu_pass_2022} pre-trains for 100 epochs on 8 A100 GPUs, which takes 60-120 hours depending on the backbone.
In addition, LUPerson, PASS, and SOLIDER each pre-train on the much larger LUPerson dataset with 4.18M images vs. our 64k images in the COCOPersons dataset.
Given the results comparison in \cref{tab:sota}, this shows our method achieves much greater pre-training efficiency for person search.
Further, we show in \cref{tab:benchmark} that SPNet-L achieves comparable metrics to other top models when using a ResNet50 backbone \cite{he_deep_2016}, with inference speed more than 5 FPS greater than the next fastest SeqNeXt model at 27.6 vs. 22.3 FPS.

\begin{table}[h]
\renewcommand{\arraystretch}{1.0}
\begin{center}
\resizebox{1.0\linewidth}{!}{
\begin{tabular}{lcccccc}
\toprule
 & & \multicolumn{2}{c}{\textbf{CUHK-SYSU}} & \multicolumn{3}{c}{\textbf{PRW}} \\
\textbf{Model} & \textbf{Backbone} & mAP & top-1 & mAP & top-1 & FPS \\
\midrule
SeqNet \cite{li_sequential_2021} & ResNet50 & 93.8 & 94.6 & 46.7 & 83.4 & 12.2 \\
SeqNeXt \cite{jaffe_gallery_2023} & ResNet50 & 93.8 & 94.3 & 51.1 & 85.8 & 22.3 \\
SPNet-L       & ResNet50 & 93.8 & 94.5 & 51.2 & 86.9 & 27.6 \\
COAT \cite{yu_cascade_2022}   & ResNet50 & 94.2 & 94.7 & 53.3 & 87.4 & 11.1 \\
\bottomrule
\end{tabular}
}
\vspace{-0.3cm}
\end{center}
\caption{Person search metrics and inference speed for models with ResNet50 backbone with ImageNet-1k classifier initialization. Inference speed in frames per second (FPS) measured on PRW with a single A100 GPU.}
\vspace{-0.5cm}
\label{tab:benchmark}
\end{table}

\begin{table}[t!]
\renewcommand{\arraystretch}{1.0}
\begin{center}
\resizebox{1.0\linewidth}{!}{
\begin{tabular}{lcccccc}
\toprule
\textbf{Config. Name} & \multicolumn{2}{c}{\textbf{Cascade Steps}} & \multicolumn{2}{c}{\textbf{Shared Heads}} & \textbf{Re-id Dim} \\
\midrule
SPNet-S & \multicolumn{2}{c}{0} & \multicolumn{2}{c}{\checkmark} & 128 \\
SPNet-L & \multicolumn{2}{c}{2} & \multicolumn{2}{c}{} & 2048 \\
\midrule
\midrule
 & \multicolumn{2}{c}{\textbf{Backbone}} & \multicolumn{2}{c}{\textbf{Post-Backbone}}  & \\
\textbf{Config. Name} & LR & WD & LR & WD &  \textbf{Image Size} \\
\midrule
Pre-train &\num{1}{-5} &    $0$ &       \num{1}{-4} &        \num{1}{-3} & 512 \\
Fine-tune &      \num{1}{-4} &    \num{5}{-4} &    \num{1}{-4} &        \num{5}{-4} & 512 or 1024\\
\bottomrule
\end{tabular}
} \vspace{-0.3cm}
\end{center}
\caption{SPNet architecture (top) and layer optimization (bottom) hyperparameter configurations. LR = Learning Rate, WD = Weight Decay.}
\vspace{-0.5cm}
\label{tab:config}
\end{table}

\subsection{Comparison with State-of-the-art}

In \cref{tab:sota} (top), we compare performance of SPNet with recent state-of-the-art person search models.
To show cases where more than one pre-training step is used, we indicate the first step with Pre-training (1) and the second with Pre-training (2).
If no pre-training is used, indicated by ``-'', all weights are randomly initialized.
We show that SPNet-L, with QC pre-training on COCOPersons, matches or exceeds other models in most metrics for the CUHK-SYSU and PRW datasets \eg, improving mAP by +1.7\% on PRW over previous SOTA LEAPS \cite{dong_leaps_2023}.

In addition, we show that the QC pre-training itself on SPNet-L adds 0.5\% mAP for CUHK-SYSU and 2.3\% mAP for PRW. This shows that the benefit from query-centric pre-training extends to the high-end of the model size / performance spectrum, even when performance statistics are nearly saturated, as on CUHK-SYSU and PRW.
In contrast, we note that OC pre-training slightly degraded performance on CUHK, demonstrating the need for SPNet and the QC pre-training approach, \ie that performance cannot be trivally improved for any person search model by simply incorporating OC pre-training.
\begin{table*}[t!]
\renewcommand{\arraystretch}{1.0}
\begin{center}
\resizebox{0.8\linewidth}{!}{
\begin{tabular}{l|c|c|c|cc|cc}
\toprule
\textbf{Person Search}       & \multirow{2}*{\textbf{Backbone}} & \textbf{Pre-training (1)} & \textbf{Pre-training (2)}    & \multicolumn{2}{c|}{\textbf{CUHK-SYSU}} & \multicolumn{2}{c}{\textbf{PRW}}  \\
   \textbf{Model}    &       & Method / Dataset & Method / Dataset & mAP   & top-1 & mAP & top-1  \\
\midrule
\multicolumn{8}{l}{\textit{SOTA Model Comparison}}  \\
\midrule
SeqNet \cite{li_sequential_2021}      & ResNet50    & Classifier / IN1k &  - & 93.8 & 94.6  & 46.7 & 83.4 \\
COAT \cite{yu_cascade_2022} & ResNet50 & Classifier / IN1k &  - & 94.2 & 94.7 & 53.3 & 87.4 \\
PSTR  \cite{cao_pstr_2022}      & PVTv2-B2 \cite{wang_pvt_2022}    &  Classifier / IN1k &  - & 95.2 & 96.2  & 56.5 & 89.7 \\
SeqNeXt \cite{jaffe_gallery_2023} & ConvNeXt-B &  Classifier / IN1k &  - & 96.1 &      96.5 & 57.6 &      89.5 \\
LEAPS \cite{dong_leaps_2023} & PVTv2-B2  &  Classifier / IN1k &  - & \textbf{96.4} & 96.9 & 59.5 & 89.7 \\
SPNet-L & ConvNeXt-B    &  Classifier / IN1k &  - & 95.9 & 96.6 & 58.9 & 89.7      \\
SPNet-L & ConvNeXt-B    &  Classifier / IN1k   &  Ours-OC / COCO & 95.8 & 96.4 & 60.7 & 90.2     \\
SPNet-L & ConvNeXt-B    &  Classifier / IN1k   &  Ours-QC / COCO & \textbf{96.4} & \textbf{97.0} & \textbf{61.2} & \textbf{90.9}     \\

\midrule

\multicolumn{8}{l}{\textit{Pre-training Comparison}}  \\
\midrule
SPNet-L & Swin-B & - & -              & 69.8 & 71.8 & 20.3 & 68.7 \\
SPNet-L & Swin-B & SOLIDER / LUP & -       & 88.0 & 89.4 & 38.1 & 81.3 \\
SPNet-L & Swin-B & SOLIDER / LUP & SOLIDER / COCO   & 88.0 & 89.4 & 38.1 & 81.3 \\
SPNet-L & Swin-B & Classifier / IN1k & - & 94.2 & 95.0 & 49.7 & 85.8 \\
SPNet-L & Swin-B & Classifier / IN1k & SOLIDER / COCO  & 94.0 & 94.8 & 49.7 & 86.1 \\
SPNet-L & Swin-B & Classifier / IN1k & Ours-OC / COCO      & 94.4 & 95.2 & 52.6 & 88.7 \\
SPNet-L & Swin-B & SOLIDER / LUP & Ours-QC / COCO      & 95.1 & 95.8 & 53.0 & 88.3 \\
SPNet-L & Swin-B & Classifier / IN1k & Ours-QC / COCO      & \textbf{95.8} & \textbf{96.3} & 54.2 & \textbf{89.0} \\
\midrule
SeqNet & Swin-B & - & - & 58.5 & 59.1 & 13.8 & 55.9 \\
SeqNet & Swin-B & Classifier / IN1k & - & 88.8 & 89.6 & 45.1 & 82.5 \\
SeqNet & Swin-B & SOLIDER / LUP     & - & 94.9 & 95.5 & \textbf{59.7} & 86.8  \\

\bottomrule
\end{tabular}
}
\vspace{-0.3cm}
\end{center}
\caption{(Top) Comparison of SOTA models. (Bottom) Comparison of performance gained from pre-training SPNet using our method vs. SOLIDER and other initialization strategies. ImageNet-1k is abbreviated as IN1k, LUPerson as LUP, and COCOPersons as COCO. When both Pre-training columns have ``-'', no pre-training is used and all weights are randomly initialized.}
\vspace{-0.4cm}
\label{tab:sota}
\end{table*}

\subsection{Comparison with Pre-training Alternatives}

In \cref{tab:sota} (bottom), we compare our pre-training approach to the backbone-only SOLIDER pre-training approach \cite{chen_beyond_2023} vs. random initialization or standard ImageNet-1k classifier initialization.
To do a fair experimental comparison, we use models with the Swin-B backbone variant from the SOLIDER codebase for all trials.
In addition, we isolate for the effect of the pre-training dataset by re-running SOLIDER on person crops from the COCOPersons dataset.
We report results from our SPNet-L model and the original SeqNet model used by SOLIDER.

For SPNet-L, we find that our QC pre-training approach outperforms all alternative initialization strategies, in some cases by wide margins \eg, +4.5\% on PRW over ImageNet-1k initialization alone.
While SOLIDER pre-training on LUPerson is significantly better than random initialization, shown in the first two rows, it is less effective than simple ImageNet-1k classifier pre-training, and far less effective than our QC pre-training approach.
In addition, when we apply additional pre-training to SOLIDER on the COCOPersons dataset, there is no improvement to fine-tuning performance, and even degradation in the case of ImageNet-1k classifier initialization.
This shows that the COCOPersons crops alone are not nearly as effective for crop-only pre-training as the LUPerson dataset, which is over $16\times$ larger.
Further, it shows these crops add no additional information beyond either LUPerson or ImageNet-1k.
This means that any benefits provided by our pre-training approach, which utilizes full scenes and not just crops, come from scene context and the pre-training method itself.

Further, we show that our QC approach exceeds our OC approach when both are initialized from ImageNet-1k classifier pre-training, with both the ConvNeXt-B backbone shown in \cref{tab:sota} (top), and the Swin-B backbone shown in \cref{tab:sota} (bottom).
This validates our reasoning that QC pre-training learns more robust features than OC pre-training.

Finally, we show that although ImageNet-1k classifier pre-training outperforms LUPerson SOLIDER pre-training for SPNet-L, SOLIDER is significantly better for the SeqNet model, as reported in the SOLIDER paper.
Differing performance for SPNet is likely due to a mismatch of the SOLIDER pre-training objective in creating effective features for the feature pyramid network in SPNet, caused by exacerbating scale misalignment of features \cite{yan_anchor-free_2021}.
In contrast, ImageNet-1k features are more general, and pair better with the feature pyramid network out-of-the-box.

\subsection{Pre-training with Noisy Labels}
\label{sec:sample_ablation}
In this section, we show that QC pre-training results in better fine-tuning performance than OC pre-training in the weakly-labeled scenario with noisy labels.
We model two noisy labels use cases: 1) ground truth annotations which should be present are missing and 2) there are additional spurious annotations.
This has applications both for manual labeling, in which persons may be missed, not all persons are annotated by design, or there is inherent labeling ambiguity as shown in \cref{fig:label_noise}.
It is also relevant for auto-labeling, in which a detector is used to label imagery, but may have low recall, creating missing annotations, or high recall but also high false positive rate, creating spurious annotations.

\begin{figure}[t]
  \centering
  \includegraphics[width=\linewidth]{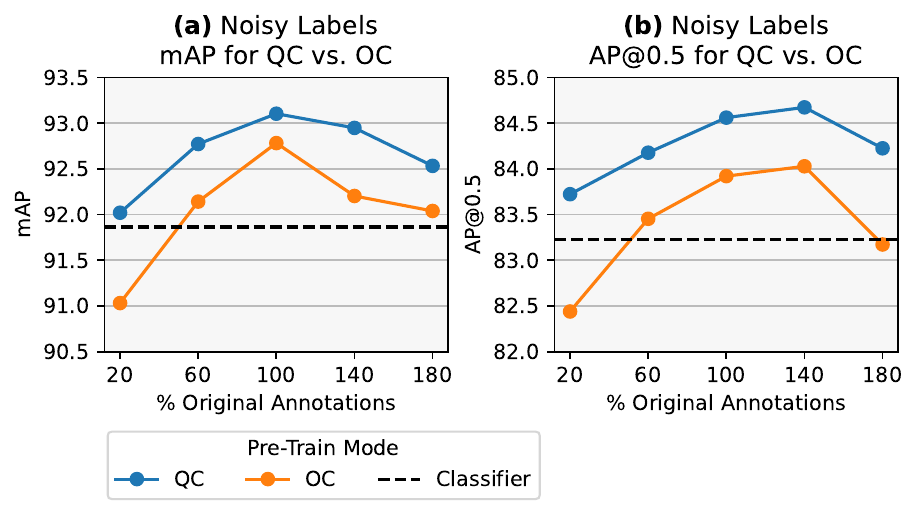}
  \vspace{-0.5cm}
    \caption{Plots of retrieval (a) and detector (b) stats for fine-tuning on the CUHK-SYSU dataset for QC and OC models pre-trained on COCOPersons with noisy labels vs. the classifier baseline.}
  \vspace{-0.5cm}
  \label{plot:noisy_labels}
\end{figure}

To model missing labels, we create successive partitions  of COCOPersons with increments of 40\% of annotations removed, with each smaller partition being a subset of all larger partitions.
To model spurious labels, we add increments of 40\% of the total labels in the original set, drawing from the existing distribution of bounding box shapes in the dataset.
The results are shown in \cref{plot:noisy_labels}, where we compare how QC and OC pre-training are affected by quantity of missing or spurious labels, as measured by fine-tuning performance on CUHK-SYSU.
Results are compared to the baseline ImageNet-1k classifier backbone initialization, shown by the black dashed line.
For all plots, we show that QC pre-training exceeds OC pre-training in all sample regimes for full fine-tuning.
Even when only 20\% of samples are used, QC pre-training offers a small benefit to fine-tuning for person search, shown by mAP in \cref{plot:noisy_labels}a, where OC pre-training actually harms fine-tuning performance by forcing the model to learn that most ground truth person boxes are background.
This trend is reflected for detection performance as well, shown in plot \cref{plot:noisy_labels}b.

\subsection{Ablation Studies}
\label{sec:ablations}

{\noindent {\bf Query-Centric vs. Object-Centric.}}
Additional ablations comparing QC vs. OC pre-training and fine-tuning are given in Supplementary \cref{supp:sec:qc_oc_ablation}. These results show that QC pre-training outperforms OC pre-training across a range of settings, including variations in the pre-training loss and variations in learning rates and weight freezing.
We also break down performance for the detection and re-id subtasks, showing that QC pre-training helps both tasks, with differences in magnitude depending on the dataset. 

{\noindent {\bf Architecture Ablations.}}
In Supplementary \cref{supp:sec:model_arch_ablation}, we show that the baseline model with FixedNorm logits achieves greater performance for all metrics over the model with BatchNorm logits.
We also consider the impact of model hyperparameters, including embedding dimension, number of cascade stages, shared vs. separate embedding heads, and backbone choice. Notably, we find that all model variants benefit from QC pre-training.

%% file: sec/5_conclusion.tex
\section{Conclusion}

We propose and validate Swap Path Net (SPNet), an end-to-end model for person search, which supports query-centric (QC) and object-centric (OC) modes of operation.
We show that the model benefits from QC pre-training and OC fine-tuning, and that pre-training can be done using only weakly-labeled person bounding boxes.
We show that 1) pre-training provides a significant boost to performance for all model variants, 2) QC pre-training benefits fine-tuning more than OC pre-training and is more robust to label noise, and 3) the model with QC pre-training achieves SOTA fine-tuning performance on CUHK-SYSU and PRW.
Finally, we show that our end-to-end person search pre-training method is more effective and efficient than backbone-only pre-training alternatives.

%% file: sec/6_acknowledgements.tex
\section*{Acknowledgements}

This work was performed under the auspices of the U.S. Department of Energy by Lawrence Livermore National Laboratory under Contract DE-AC52-07NA27344, with release number LLNL-CONF-2001396.

%% file: sec/X_suppl.tex
\clearpage
\setcounter{page}{1}
\maketitlesupplementary

\textit{This supplementary material contains content beneficial to but not required for understanding of the main paper.
It includes information about code used to produce results from the paper, additional dataset metadata, implementation details, and results of additional ablations which explore the hyperparameter design space.}

\section{Code and Reproduction}
We include all code, configs, and instructions required to reproduce results from the paper in the corresponding GitHub repository\footnote{Project repository: \url{https://github.com/LLNL/spnet}}.

\section{Dataset Metadata}
\label{supp:sec:metadata}

The two standard person search benchmark datasets we use for model fine-tuning are the CUHK-SYSU dataset \cite{xiao_joint_2017} and the PRW dataset \cite{zheng_person_2017}.

The CUHK-SYSU dataset has 18k scenes with 95k annotations, split among 23k boxes for 8.5k known identities, and 72k boxes for unknown identities. The scenes in CUHK-SYSU come from a mixture of handheld device photos taken on city streets, and scenes from TV shows and films. The standard test retrieval scenario for CUHK-SYSU uses 2,900 queries with 100 scenes in each gallery.

The PRW dataset has 12k scenes with 43k annotations, split among 34k boxes for 1k known identities, and 9k boxes for unknown identities. The scenes from PRW come from six fixed cameras installed at the Tsinghua University campus in Beijing. The standard test retrieval scenario for PRW uses 2,057 queries with all 6,112 test scenes in the gallery.

\section{Additional Implementation Details}
\label{supp:sec:imp_details}

{\noindent {\bf Model Configurations.}}
In all model versions, a 4-layer MLP is applied to offset embeddings $x_o$ to produce box regression offsets, while classification logits are computed directly from $\| x_o \|$. The training batch size is 8, with $k=2$ augmentations per image during pre-training.

{\noindent {\bf Anchor Sampling.}} To reduce differences between QC and OC pre-training from anchor sampling during detection, we ensure that the same number of anchors are optimized in each batch between QC and OC trials (2,048 by default).
For the OC case, this is calculated as number of images per batch $\times$ number of anchors per image.
For the QC case, this is calculated as number of images per batch $\times$ number of query-image pairs $\times$ number of anchors per query-image pair.

Query-image pairs constitute the pairing of a given query embedding $x_q$ with anchor embeddings from a given image $x_a$. With the total number of query-image pairs equal to number of queries $\times$ number of images, this can quickly exceed memory limitations if all pairs are used, so some subsampling procedure is required. We select pairs in the following order up to a fixed number (32 by default): 1) queries are selected for images which they are not present in, but share a label with a box in the image, 2) queries are selected for images which they are present in, and 3) queries are selected for images which they are neither present in, nor share a label with any boxes in.

Both QC and OC sampling methods attempt to balance the number of positive and negative samples, but there are usually more negative samples in practice due to sparsity of positives and typical hyperparameter selection.

{\noindent {\bf Image Augmentation.}}
We adopt three methods of image augmentation, which all consist of combinations of scaling, cropping, and horizontal flipping. For consistency, we label these here using the configuration name used in the code. We call the standard augmentation for person search \textit{window resize} as in \cite{jaffe_gallery_2023}, abbreviated \texttt{wrs}. This consists of scaling the image to fit in a window with shortest side length $900$ and longest side length $1,500$. \texttt{wrs} augmentation is used only for evaluation. In \texttt{rrc2} augmentation, we first perform \texttt{wrs} scaling, then randomly select from two types of crops (followed by random horizontal flip): 1) fixed size square random crop containing at least one bounding box in the image, chosen at random, and 2) random sized crop containing all bounding boxes in the image, that is then resized to a fixed square size. These square crop sizes are $512 \times 512$ during pre-training and SPNet-S fine-tuning, and $1024 \times 1024$ during SPNet-L fine-tuning. Crop size is one of the most important parameters for controlling memory usage. \texttt{rrc2} augmentation is used for all fine-tuning runs. \texttt{rrc\_scale} augmentation is the same as \texttt{rrc2} augmentation, except we scale randomly in the range $0.5 \times$ to $2 \times$ instead of \texttt{wrs} scaling. \texttt{rrc\_scale} augmentation is used for all pre-training runs.

{\noindent {\bf Training Times.}}
All models were trained using a single A100 GPU with 82GB VRAM.
Pre-training and fine-tuning times are shown in \cref{tab:training_time}, with the QC pre-training taking 30 hours for SPNet-S and 46.5 hours for SPNet-L.
We note that OC pre-training is more efficient, taking about 6 hours less for either model variant.

\begin{table*}
\renewcommand{\arraystretch}{1.0}
\begin{center}
\begin{tabular}{lccccc}
\toprule
 & & \textbf{Pre-train (QC)} & \textbf{Pre-train (OC)} & \multicolumn{2}{c}{\textbf{Fine-tune (OC)}} \\
\textbf{Model} & \textbf{Backbone} & COCOPersons & COCOPersons & CUHK-SYSU & PRW \\
\midrule
SPNet-S & ConvNeXt-T & 30.0h & 24.5h & 2.5h & 1.5h \\
SPNet-L & ConvNeXt-B & 46.5h & 40.5h & 9.0h & 4.5h \\
\bottomrule
\end{tabular}
\end{center}
\caption{Training times for person search model variants on a single A100 GPU. Pre-training and fine-tuning are each done for 30 epochs on the respective datasets.}
\label{tab:training_time}
\end{table*}

\section{Supporting Ablations}

\subsection{QC vs. OC Ablations}
\label{supp:sec:qc_oc_ablation}

{\noindent {\bf Subtask Performance.}}
In \cref{tab:metric_breakdown}, we compare QC vs. OC pre-training to ImageNet-1k Classifier pre-training, and random backbone initialization.
The comparison is done for SPNet-S with a ConvNeXt-Tiny backbone, and we show metrics of person search (mAP) in addition to metrics of re-id (GT mAP) and detection (AP@0.5).
This helps us understand the contribution of pre-training to detection, re-id, and the combined person search problem.
Measuring ground truth re-id performance (GT mAP) is equivalent to using a perfect object detector, and helps us understand re-id performance independent of detector quality.

\begin{table*}
\renewcommand{\arraystretch}{1.0}
\begin{center}
\begin{tabular}{lcccccc}
\toprule
 & \multicolumn{3}{c}{\textbf{CUHK-SYSU}} & \multicolumn{3}{c}{\textbf{PRW}} \\
\textbf{Pre-train Method} & mAP & GT mAP & AP@0.5 & mAP & GT mAP & AP@0.5 \\
\midrule
Random Init. & 69.8 & 85.8 & 47.8 & 22.3 & 28.7 & 66.8 \\
Classifier & 91.9 & 95.8 & 83.2 & 52.7 & 56.6 & 88.8 \\
Ours-OC & 92.8 & \textbf{96.3} & 83.9 & 54.5 & 57.6 & \textbf{89.1} \\
Ours-QC & \textbf{93.3} & 95.9 & \textbf{84.9} & \textbf{55.6} & \textbf{58.7} & 88.9 \\
\bottomrule
\end{tabular}
\end{center}
\caption{Person search (mAP), re-id (GT mAP), and detection metrics (AP@0.5) for pre-training method comparison on SPNet-S with ConvNeXt-Tiny backbone.}
\label{tab:metric_breakdown}
\end{table*}

We show both QC and OC pre-training improve over ImageNet-1k Classifier pre-training for all metrics, and that QC is superior to OC pre-training for most metrics.
Most importantly, we show that QC pre-training exceeds OC pre-training for person search (mAP) on both datasets (+0.5\% on CUHK-SYSU, +1.1\% on PRW), though whether re-id or detection benefit more depends on the dataset. Finally, we show that all pre-training vastly exceeds random initialization. Additional ablations comparing QC vs. OC pre-training and fine-tuning are shown in the following subsections. These results show that QC pre-training outperforms OC pre-training across a range of different conditions.

{\noindent {\bf Re-id Pre-training.}}
In \cref{tab:reid_ablation}, we compare different methods of handling the re-id loss during pre-training. The goal was to isolate the effect of the re-id loss on pre-training, for both QC and OC methods. We found that pre-training only the re-id loss, without optimizing the other losses, was not beneficial for all statistics. 

\begin{table}
\begin{center}
\begin{tabular}{lcccc}
\toprule
 & \multicolumn{2}{c}{\textbf{CUHK-SYSU}} & \multicolumn{2}{c}{\textbf{PRW}}  \\
\textbf{Method} & mAP & top-1 & mAP & top-1 \\
\midrule
Baseline & 91.9 & 93.2 & 52.7 & 86.6 \\
Re-id Only & 91.5 & 93.0 & 51.9 & 87.4 \\
OC Re-id GT & 92.4 & 93.7 & \textbf{55.7} & 89.1 \\
QC Re-id GT & 93.2 & \textbf{94.3} & 55.1 & 88.5 \\
OC Original & 92.8 & 94.1 & 54.5 & 88.8 \\
QC Original & \textbf{93.3} & 94.1 & 55.6 & \textbf{89.5} \\
\bottomrule
\end{tabular}
\end{center}
\caption{Comparison of different settings of the re-id pre-training loss for fine-tuning performance. For \textit{re-id GT}, only GT box embeddings are used to compute re-id loss, and for \textit{re-id only}, we optimize only the re-id loss (with GT boxes) and no other losses. In \text{original} trials, both GT and detected box embeddings are used to compute re-id loss.}
\label{tab:reid_ablation}
\end{table}

We found that QC pre-training using detected boxes with IoU $\geq 0.7$ performed best on balance. When only ground truth (GT) boxes were used to compute the re-id loss, and no detected boxes, we found that QC pre-training was impaired more than OC pre-training.

\begin{figure*}
  \centering
  \includegraphics[width=0.65\linewidth]{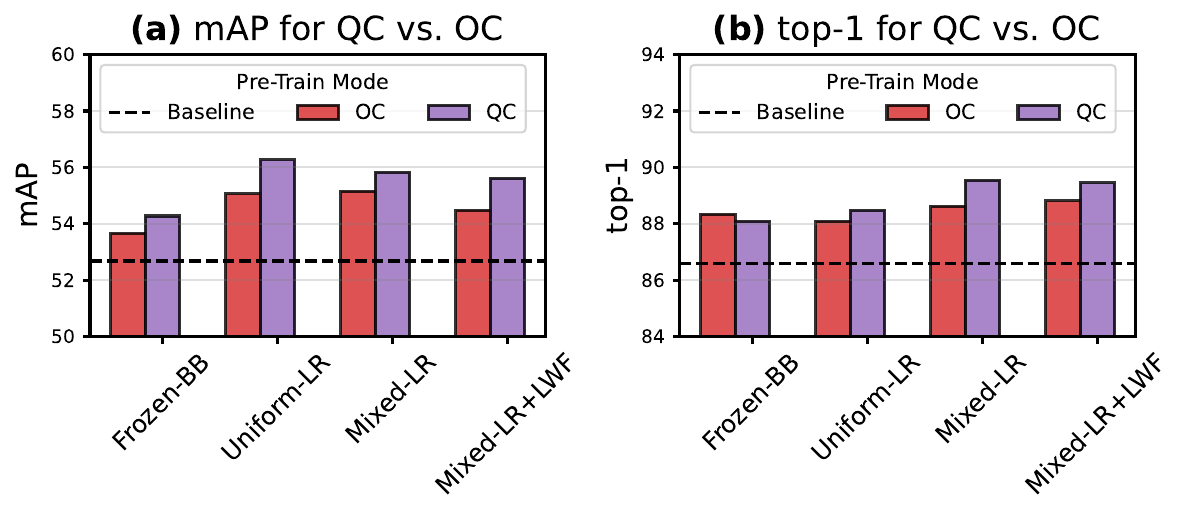}
  \caption{Comparison of SPNet fine-tune performance on PRW with QC vs. OC pre-training for varying layer optimization hyperparameters (configurations described in \cref{tab:config_full}).}
  \label{plot:prw_backbone_ablation}
\end{figure*}

{\noindent {\bf Query-Centric vs. Object-Centric Pre-training.}}
We compare query-centric vs. object-centric pre-training for four configurations, given in \cref{tab:config_full} (bottom), with results in \cref{plot:prw_backbone_ablation} for the PRW dataset. All models with pre-training significantly outperform the baseline in mAP and top-1 accuracy, and nearly all QC models outperform the corresponding OC model. One exception is the Frozen Backbone model for top-1 accuracy, showing that the query-centric pre-training performs better when the backbone is optimized in addition to later layers.

\begin{table*}
\begin{center}
\begin{tabular}{llccccc}
\toprule
\textbf{Config. Name} &\textbf{Ablation Name} & \multicolumn{2}{c}{\textbf{Cascade Steps}} & \multicolumn{2}{c}{\textbf{Shared Heads}} & \textbf{Re-id Dim} \\
\midrule
SPNet-S & c0-share-d128 & \multicolumn{2}{c}{0} & \multicolumn{2}{c}{\checkmark} & 128 \\
- & c0-sep-d128 & \multicolumn{2}{c}{0} & \multicolumn{2}{c}{} & 128 \\
- & c2-share-d128 & \multicolumn{2}{c}{2} & \multicolumn{2}{c}{\checkmark} & 128 \\
- & c2-sep-d128 & \multicolumn{2}{c}{2} &\multicolumn{2}{c}{}  & 128 \\
SPNet-L & c2-sep-d2048 & \multicolumn{2}{c}{2} & \multicolumn{2}{c}{} & 2048 \\
\midrule
\midrule
 & \multicolumn{2}{c}{\textbf{Backbone}} & \multicolumn{2}{c}{\textbf{Post-Backbone}}  & \\
\textbf{Config. Name} &\textbf{Ablation Name} & LR & WD & LR & WD & \textbf{LwF Loss} \\
\midrule
- & Uniform LR &      \num{1}{-4} &    \num{1}{-3} &    \num{1}{-4} &        \num{1}{-3} & \\
Pre-train & Mixed LR &      \num{1}{-5}  &   $0$ &       \num{1}{-4} &        \num{1}{-3} &  \\
- & Mixed LR + LwF &\num{1}{-5} &    $0$ &       \num{1}{-4} &        \num{1}{-3} & \checkmark \\
- & Frozen-BB &     Frozen &        Frozen &        \num{1}{-4} &        \num{1}{-3} & \\
Fine-tune & Fine-Tune &      \num{1}{-4} &    \num{5}{-4} &    \num{1}{-4} &        \num{5}{-4} & \\
\bottomrule
\end{tabular}
\end{center}
\caption{SPNet architecture (top) and layer optimization (bottom) hyperparameter configurations. LR = Learning Rate, WD = Weight Decay, LwF = Learning without Forgetting \cite{li_learning_2018}.}
\label{tab:config_full}
\end{table*}

For pre-training, we also explore the effect of the \textit{Learning without Forgetting} (LwF) loss from \cite{li_learning_2018}, equivalent to the knowledge distillation loss from \cite{hinton_distilling_2015}. Since we initialize our model backbone from ImageNet-1k classifier weights before pre-training, the idea is that this loss will help preserve useful features learned during the classifier pre-training. This loss $\mathcal{L}_\text{\tiny{LwF}}$ is simply added to the other losses when used, using temperature $T=2$ as in \cite{li_learning_2018}.
As shown in \cref{plot:prw_backbone_ablation}, we did not find use of the LwF loss to result in consistently better performance, so it was not used for other trials.

{\noindent {\bf Query-Centric vs. Object-Centric Fine-tuning.}}
While we primarily explore OC fine-tuning (OC-FT) and evaluation in this work, our framework supports QC fine-tuning (QC-FT) and evaluation as well. QC fine-tuning is done using the same procedure as QC pre-training, with $k=2$ augmentations per image, but using the \texttt{rrc2} augmentation method.

In \cref{tab:qc_oc_baseline}, we compare QC-FT and OC-FT  for models with and without COCOPersons QC pre-training.
While the QC-FT model outperforms the OC-FT model on PRW top-1 (without PT), OC-FT performance is drastically higher for mAP and top-1 on CUHK-SYSU ($>$10\% mAP).
In addition, the OC-FT model benefits significantly from QC pre-training for both datasets and metrics, while the QC-FT model benefits only slightly on PRW and is actually harmed on CUHK-SYSU.

\begin{table}
\begin{center}
\begin{tabular}{lcccc}
\toprule
 & \multicolumn{2}{c}{\textbf{CUHK-SYSU}} & \multicolumn{2}{c}{\textbf{PRW}}  \\
\textbf{Method} & mAP & top-1 & mAP & top-1 \\
\midrule
OC-FT & 91.9 & 93.2 & 52.7 & 86.6 \\
OC-FT+PT & 93.3 & 94.1 & 55.6 & 89.5 \\
\textit{Gain from PT} & \small{\textcolor{LimeGreen}{+1.4}} & \small{\textcolor{LimeGreen}{+0.9}} & \small{\textcolor{LimeGreen}{+2.9}} & \small{\textcolor{LimeGreen}{+2.9}} \\
\midrule
\midrule
QC-FT & 80.8 & 87.7 & 54.0 & 86.7 \\
QC-FT+PT & 77.4 & 85.4 & 54.1 & 87.1 \\
\textit{Gain from PT} & \small{\textcolor{red}{-3.4}} & \small{\textcolor{red}{-2.3}} & \small{\textcolor{LimeGreen}{+0.1}} & \small{\textcolor{LimeGreen}{+0.4}} \\
\bottomrule
\end{tabular}
\end{center}
\caption{Comparison of QC and OC fine-tuning (-FT) for baseline model vs. COCOPersons QC pre-trained model (+PT).}
\label{tab:qc_oc_baseline}
\end{table}

This suggests QC fine-tuning is more impacted by the distribution shift between pre-training and fine-tuning datasets, at least for the shared embedding head model. It also suggests the QC pre-train to OC fine-tune transition has a regularizing effect, limiting the effect of overfitting on the pre-training dataset.  
\subsection{Model Architecture Ablation}
\label{supp:sec:model_arch_ablation}
\begin{figure*}
  \centering
  \includegraphics[width=0.65\linewidth]{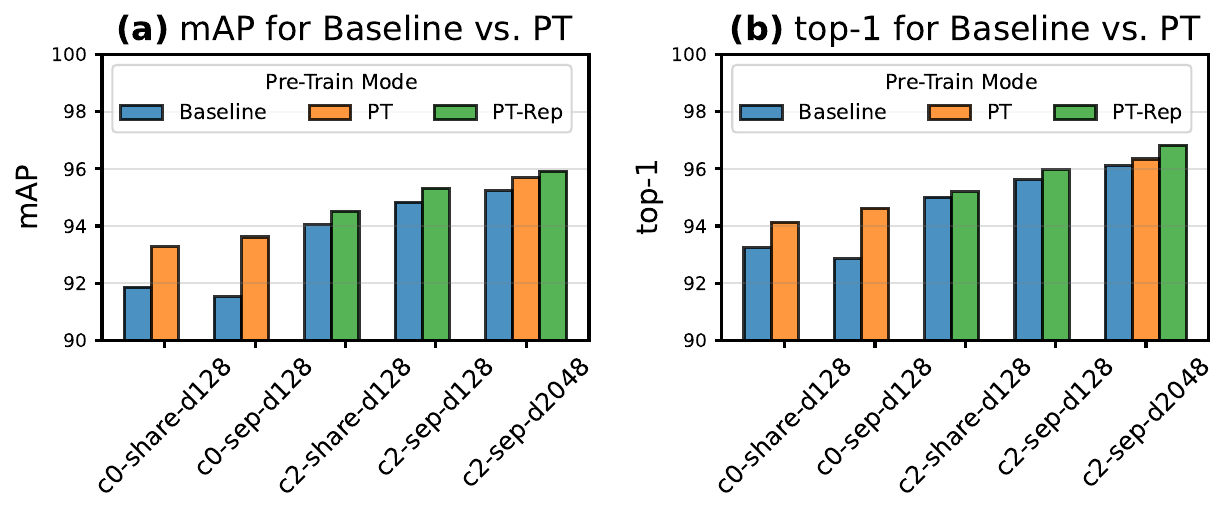}
  \caption{Comparison of SPNet fine-tune performance on CUHK-SYSU with and without QC pre-training for varying architecture hyperparameters (configurations described in \cref{tab:config_full}). PT=pre-trained, PT-Rep=pre-trained with replicated weight loading for cascade layers.}
  \label{plot:cuhk_arch_ablation}
\end{figure*}

{\noindent {\bf Model Architecture.}}
To understand the model architecture design space, we examine variations in number of cascade steps, shared vs. separate embeddings heads, and size of the re-id embedding dimension. In \cref{plot:cuhk_arch_ablation}, we compare the configurations described in \cref{tab:config_full} (top). We show that the cascaded model with two steps is better than the base model, using separate embedding and re-id heads results in better fine-tuning performance, and larger re-id embedding dimension is beneficial. In addition, all architecture configurations benefit from pre-training, although the cascaded models benefit less. 

{\noindent {\bf Cascaded Weight Loading.}}
When loading weights from pre-trained SPNet into a larger version with multiple cascade stages, we have to make a choice about how to load weights into layers not utilized during pre-training. We explore two simple options: loading weights only into layers that were pre-trained, and duplicating weights into equivalent layers in each cascade stage, shown by \textit{PT} vs. \textit{PT-Rep} in \cref{plot:cuhk_arch_ablation}. For the \texttt{c2-sep-d2048} configuration, the PT-Rep method of weight loading outperforms PT, showing that the benefits of the pre-trained weight initialization extend to cascade stages, even though the pre-trained model was trained without cascading.

{\noindent {\bf Classifier Logits.}}
Recall from \cref{sec:losses} that the offset embedding is defined as the difference between query and anchor embeddings: $x_o = x_q - x_a$ with $x_o \in \mathbb{R}^d$ and $w = \| x_o \|$.
Then the class logits $z$ can be calculated as

\begin{equation}
\label{eq:z_logits}
z = \frac{w - \mu_\text{Chi}}{s_\text{Chi}}
\end{equation}

with

\begin{equation}
\mu = \sqrt{2} \frac{
    \Gamma \left( \frac{d+1}{2} \right) }{ 
    \Gamma \left( \frac{d}{2} \right) },
 \quad s = \sqrt{d - \mu^2} 
\end{equation}

We note for $d \gg 1$, $\mu \approx \sqrt{d - \frac{1}{2}}$ and $s \approx \sqrt{\frac{1}{2}}$.

The \textit{FixedNorm} transformation in \cref{eq:z_logits} has two key properties:
1) For $x_o \sim \mathcal{N}(0, I_d)$, $\mathbb{E}[z] = 0, \mathrm{Var}[z] = 1$, independent of $d$. In addition, $\lim_{{d \to \infty}} z \sim \mathcal{N}(0,1)$ due to the Central Limit Theorem. Although the distribution of $x_o$ is rarely unit-normal and changes during optimization, this framing gives us a reasonable choice of shifting and scaling parameters which works well in practice, does not require learnable parameters or hyperparameters, and holds independent of $d$.

2) For $x_q$ similar to $x_a$ \ie, $\|x_o\|$ small, $z$ is larger, and for $x_q$ dissimilar to $x_a$ \ie, $\|x_o\|$ large, $z$ is smaller. Intuitively, when a query embedding matches a given anchor embedding, according to box IoU overlap, we want the two to be more similar, and when they do not match, we want them to be more different. The relationship is also critical to co-optimizing with the re-id loss on $x_q$ and has a nice relationship with the box regression loss.

To validate our FixedNorm logits over the BatchNorm logits from \cite{chen_norm-aware_2020}, we compare both for the baseline model, with results shown in \cref{tab:norm_logits}. The FixedNorm method achieves greater performance for all metrics, especially on the PRW dataset, where mAP exceeds the BatchNorm method by more than 10\%. This is likely due in part to unbalanced batch statistics, which are dominated by negative samples, unlike the original Norm-Aware Embedding use case.
\cref{tab:norm_logits} also shows that negating the norm after scaling is beneficial, especially for the FixedNorm logits, validating the rationale discussed above.

\begin{table}
\begin{center}
\begin{tabular}{lcccc}
\toprule
 & \multicolumn{2}{c}{\textbf{CUHK-SYSU}} & \multicolumn{2}{c}{\textbf{PRW}}  \\
\textbf{Method} & mAP & top-1 & mAP & top-1 \\
\midrule
BatchNorm Logits (+) & 90.4 & 91.9 & 49.8 & 84.3 \\
BatchNorm Logits (-) & 91.2 & 92.0 & 42.2 & 82.2 \\
FixedNorm Logits (+) & 91.1 & 92.2 & 52.6 & 86.3 \\
FixedNorm Logits (-) & \textbf{91.9} & \textbf{93.2} & \textbf{52.7} & \textbf{86.6} \\
\bottomrule
\end{tabular}
\end{center}
\caption{Comparison of BatchNorm Logits from \cite{chen_norm-aware_2020} to proposed FixedNorm Logits for CUHK-SYSU and PRW datasets, for the baseline model. (+) vs. (-) indicates whether positive or negative embedding norm was used to compute logits.}
\label{tab:norm_logits}
\end{table}